\documentclass[10pt,twocolumn,letterpaper]{article}
\usepackage[pagenumbers]{cvpr}

\usepackage{xspace}
\newcommand{\methodname}{GDRO}
\newcommand{\method}{\texttt{\methodname}\xspace}

\usepackage{algorithm}
\usepackage{algpseudocode}
\usepackage{float}
\usepackage{subcaption}
\usepackage{stfloats}

\definecolor{cvprblue}{rgb}{0.21,0.49,0.74}
\usepackage[pagebackref,breaklinks,colorlinks,allcolors=cvprblue]{hyperref}

\title{\method: Group-level Reward Post-training Suitable for Diffusion Models}

\author{
    Yiyang Wang\textsuperscript{1}\quad
    Xi Chen\textsuperscript{1}\quad
    Xiaogang Xu\textsuperscript{2}\quad
    Yu Liu\textsuperscript{3}\quad
    Hengshuang Zhao\textsuperscript{$1,\dagger$}\\[2pt]
    {\textsuperscript{1}The University of Hong Kong \quad
    \textsuperscript{2}The Chinese University of Hong Kong \quad
    \textsuperscript{3}Tongyi Lab
    } \\[0.3em]
}

\newcommand\nnfootnote[1]{%
  \begin{NoHyper}
  \renewcommand\thefootnote{}\footnote{#1}%
  \addtocounter{footnote}{-1}%
  \end{NoHyper}
}

\begin{document}
\twocolumn[{%
\renewcommand\twocolumn[1][]{#1}%
\maketitle
\begin{center}
    \centering
    \captionsetup{type=figure}
    \vspace{-10pt} 
    \includegraphics[width=\textwidth]{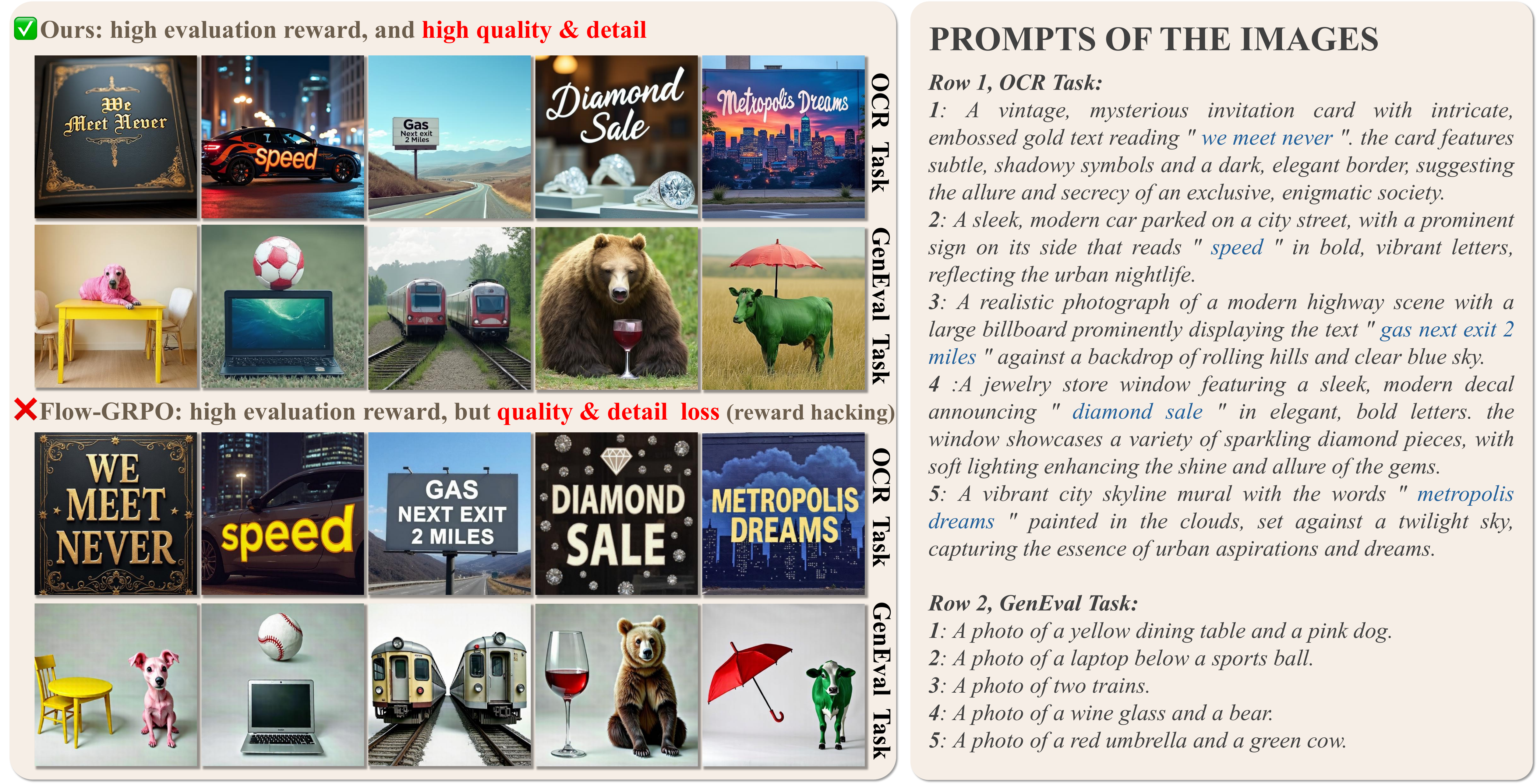}
    \vspace{-15pt}
    \captionof{figure}{
        \textbf{Illustrations of our method}. As shown in the top block, \method optimizes the text-to-image model toward high precision OCR rendering and layout planning, while maintaining high quality and prompt-image alignment.
        In contrast, in the bottom block, although achieving a high evaluation score, Flow-GRPO hacks the reward to make the text very big and the content of the image unnatural.  
        }
    \label{fig: teaser}
\end{center}
}]

\nnfootnote{
\hspace{-2em}$\dagger$~{Corresponding Author.}
}

\begin{abstract}
   Recent advancements adopt online reinforcement learning (RL) from LLMs to text-to-image rectified flow diffusion models for reward alignment. The use of group-level rewards successfully aligns the model with the targeted reward. However, it faces challenges including low efficiency, dependency on stochastic samplers, and reward hacking. The problem is that rectified flow models are fundamentally different from LLMs: 1) For efficiency, online image sampling takes much more time and dominates the time of training. 2) For stochasticity, rectified flow is deterministic once the initial noise is fixed. Aiming at these problems and inspired by the effects of group-level rewards from LLMs, we design Group-level Direct Reward Optimization (\method). \method is a new post-training paradigm for group-level reward alignment that combines the characteristics of rectified flow models. Through rigorous theoretical analysis, we point out that \method supports full offline training that saves the large time cost for image rollout sampling. Also, it is diffusion-sampler-independent, which eliminates the need for the ODE-to-SDE approximation to obtain stochasticity. We also empirically study the reward hacking trap that may mislead the evaluation, and involve this factor in the evaluation using a corrected score that not only considers the original evaluation reward but also the trend of reward hacking. Extensive experiments demonstrate that \method effectively and efficiently improves the reward score of the diffusion model through group-wise offline optimization across the OCR and GenEval tasks, while demonstrating strong stability and robustness in mitigating reward hacking.
    
\end{abstract}
\section{Introduction}
\label{sec:intro}

Post-training using group-level rewards has been widely employed in LLMs, aiming at maximizing certain types of rewards in a group-wise manner~\cite{guo2025deepseek}.
Such great success in LLMs inspires studies that bring group-level rewards to diffusion models~\cite{ho2020denoising, Rombach_2022_CVPR}.
These methods use online reinforcement learning (RL)~\cite{jaques2017sequence, kreutzer2018reliability}, where the diffusion model repeatedly synthesizes online samples (rollouts) to assign rewards at each optimization step, and gets PPO-style~\cite{schulman2017proximal} or GRPO-style~\cite{guo2025deepseek} updates with the policy gradient~\cite{black2024training,fan2023reinforcement}.
However, this framework requires stochasticity on each diffusion step that contradicts the deterministic sampler of rectified flow models~\cite{liu2022flow}. Recent works~\cite{liu2022flow,xue2025dancegrpo} overcome this constraint by approximating the ODE sampler with an SDE sampler.

Despite these advancements, the online RL algorithms face several issues with diffusion models. 
1) Low efficiency: 
The diffusion model has to calculate the full diffusion chain to sample images.
But the online RL algorithms require such online sampling processes across optimization steps, making online rollout sampling dominate the training and cost significantly.
2) Sampler dependency:
These online RL algorithms require stochastic diffusion samplers to calculate policy gradients for optimization.
But rectified flow models are inherently deterministic once the initial noise is fixed.
Though Flow-GRPO overcomes this constraint by approximating the native ODE sampler with SDE,
such an approximation may bring out-of-domain issues and quality degradation.
3) Reward hacking:
Though RL methods successfully optimize the model toward higher rewards, reward hacking causes severe degradation in quality, detail, and prompt-image alignment.

To address the above problems and leverage the advantages of group-level rewards from LLMs, we design group-level direct reward optimization (\method) that combines the characteristics of diffusion models.
\method is a post-training algorithm for group-wise reward alignment on diffusion models. 
Through rigorous theoretical analysis, we prove that \method can perform group-level reward optimization by manipulating the implicit reward functions that can be computed at any diffusion timestep.
This implicit reward function has a solid theoretical background from DPO-based methods~\cite{rafailov2023direct,wallace2024diffusion}, only requiring the noise or velocity predicted by the diffusion model.
With this guarantee, \method bypasses the need for log probability computations across the diffusion timesteps that require stochasticity in the diffusion sampler, making it a sampler-independent method that eliminates the need for ODE-to-SDE approximation. 
What's more, because the implicit reward function can be computed at any diffusion timestep, \method doesn't have to compute the whole diffusion sampling chain starting from Gaussian noise, making it able to operate in a totally offline way, greatly saving the costs for online sampling.

For reward hacking, we also perform an in-depth study on it and find that we can use automatic metrics to reflect the trend of reward hacking. Based on this finding, we propose a corrected score, which combines the original reward model with the reward hacking trend and serves as a more objective metric for evaluation.
Extensive experiments show that \method successfully optimizes the model toward higher evaluation rewards efficiently on the OCR and the GenEval task~\cite{ghosh2023geneval}. Combined with visualizations and the corrected score, we also find that \method effectively mitigates the reward hacking issue, demonstrating its robustness in quality preservation compared to RL methods.

To summarize, this paper proposes \method, a post-training algorithm for group-level reward optimization designed for diffusion models. 
\method supports full offline training and is sampler-independent.
We also analyze the phenomenon of reward hacking and propose a corrected score to evaluate the performance while considering reward hacking.
With this corrected score, experiments demonstrate \method efficiently optimizes the diffusion model toward higher rewards, while effectively mitigating the reward hacking.    
\section{Related Work}
\label{sec:related}
\noindent \textbf{Reward post-training for LLMs.}
Online RL has been widely adopted for LLMs in RLHF~\cite{ouyang2022training, achiam2023gpt} or RLVR~\cite{guo2025deepseek}.
These methods update the model to maximize a reward or follow some preferences using policy gradient algorithms such as PPO~\cite{schulman2017proximal} or GRPO~\cite{shao2024deepseekmath}.
Apart from online RL, DPO~\cite{rafailov2023direct} reformulates the online RL objective to support offline pairwise optimizations.
The success of these methods in LLMs brings inspiration to visual content generation. 

\noindent \textbf{Reward post-training for vision generation.}
AlignProp~\cite{clark2024directly, prabhudesai2023aligning} and DiffDoctor~\cite{wang2025diffdoctor} directly optimize the diffusion model by directly maximizing and backpropagating from a differentiable reward function. However, they don't support reward functions that are not differentiable.
DDPO~\cite{black2024training} and DPOK~\cite{fan2023reinforcement} apply online RL to diffusion models by considering the diffusion process as an MDP process. But they only work on the DDPM-based diffusion models~\cite{ho2020denoising, song2020denoising, song2021scorebased} with stochastic samplers and therefore do not support rectified flow diffusion models~\cite{liu2022flow}.
Flow-GRPO~\cite{liu2025flow} and DanceGRPO~\cite{xue2025dancegrpo} approximate the ODE process of rectified flow models using an SDE sampler, thereby enabling training rectified flow diffusion models using online RL.
Diffusion-DPO~\cite{wallace2024diffusion} approximates the original DPO objective in the context of diffusion models.
However, similar to DPO, it only supports pair-wise optimizations and cannot leverage the explicit reward information.
In this paper, we propose \method that doesn't have the requirement for differentiable reward functions or the need for a stochastic diffusion sampler during training. It also supports group-wise training with explicit rewards involved rather than solely pair-wise preference training.

\section{Method}
Previous RL algorithms for diffusion models cost much on online sampling and require ODE-to-SDE approximation.
Therefore, we aim to bypass the online sampling and stochasticity dependency in our derivation toward \method, starting from the implicit reward function.

\subsection{Preliminaries}
\label{sec: Preliminaries}
\noindent \textbf{RL fine-tuning.}
RL fine-tuning~\cite{jaques2017sequence, kreutzer2018reliability} aims to optimize a conditional policy model $\pi_\theta(x_0|c)$ 
to maximize a reward model $r(x_0, c)$ (\textit{e.g.,} OCR text accuracy on text-to-image models) that is defined on its outputs ($x_0$) (\textit{e.g.,} image) and conditions ($c$) (\textit{e.g.,} prompt), while being regularized on a KL-divergence from a reference policy model $\pi_\text{ref}$:
\begin{align}
\label{formulation: RL}
    \max_{\theta}\mathbb{E}_{c,x_0\sim\pi_\theta}r(x_0,c)-\beta_\text{KL}\mathbb{D}_\text{KL}\big[\pi_\theta(x_0|c) || \pi_\text{ref}(x_0|c) \big],
\end{align}
where $\beta_\text{KL}$ controls the KL regularization strength.

\noindent \textbf{Implicit reward function.}
DPO~\cite{rafailov2023direct} reformulates the RL fine-tuning objective above by reparameterizing the reward model term $r(\cdot)$ into an implicit reward function. Specifically, the optimal target policy $\pi_\theta^*$ of \cref{formulation: RL} is
\begin{equation}
    \pi_\theta^*(x_0|c) = \frac{\pi_\text{ref}(x_0|c)e^{r(x_0,c)/\beta_\text{KL}}}{Z(c)},
\end{equation}
where $Z(c) = \sum_{x_0} \big[\pi_\text{ref}(x_0|c)e^{r(x_0,c)/\beta_\text{KL}} \big] $ is the partition function. 
With this, the reward function is reformulated as
\begin{equation}
\label{formulation: naive_implicit_reward}
    s_\theta(x) = r(x_0, c) = \beta_\text{KL} \log \frac{\pi_\theta^*(x_0\mid c)}{\pi_\text{ref}(x_0\mid c)}
+ \underbrace{\beta_\text{KL} \log Z(c)}_{\text{can be deleted}}.
\end{equation}
We call this the implicit reward function calculated from the policy model, and the DPO objective can be formulated with this implicit reward function:
\begin{equation}
\label{formulation: DPO_objective}
    \mathcal{L}_\text{DPO}(\theta) = -\log\sigma\big(s_\theta(x_0^+, c) -s_\theta(x_0^-,c) \big),
\end{equation}
where $x_0^+$ is the chosen sample and $x_0^-$ is the rejected sample.
The scalar term $\beta \log Z(c)$ can be deleted in \cref{formulation: naive_implicit_reward} because it's eliminated in the calculation.

Diffusion-DPO~\cite{wallace2024diffusion} proves that the implicit reward function can be approximated with the following equation in the context of diffusion models at a given diffusion timestep $t$:
\begin{align}
\label{formulation: diffusion_implicit_reward}
s_\theta(x,t) =
 -\beta\,\mathbb{E}_{t,v}\!\big[\|v-v_\theta(C)\|_2^2
 - \|v-v_{\mathrm{ref}}(C)\|_2^2\big],
\end{align}
where 
$C=\big(x_t(c),t,c\big)$, 
$v$ is the perturb noise and $x_t(c)$ is the perturbed image at timestep $t$ of the rectified flow models~\cite{liu2022flow}, 
$\beta=T\beta_\text{KL}$ ($T$: total timestep number),
and $v_\theta(C)$ is the diffusion model output.
The above derivations show that the implicit reward function only requires the perturbed images, making no online sampling or stochasticity involved for DPO.
We will also leverage this implicit reward function in our following derivations toward \method.

\begin{figure}[t]
  \centering
  \resizebox{!}{0.48\linewidth}{
  \includegraphics{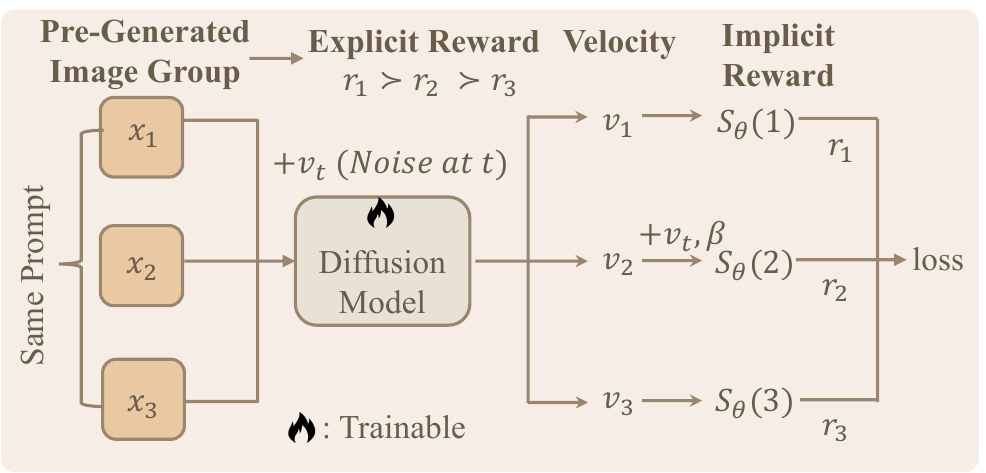}
  }
  \vspace{-15pt}
  \caption{\textbf{Overview of our method.} Given a pre-generated image group synthesized from the same prompt and their corresponding explicit rewards, we perturb the images with noise on different time steps, feed them to the diffusion model to predict the velocity, and calculate the implicit rewards accordingly to get the final loss. 
  }
  \vspace{-10pt}
  \label{fig: pipeline}
\end{figure}

\subsection{Group-level Direct Reward Optimization}
\label{sec: gdpo}
We then consider group-level rewards that bring great success to LLM post-training.
With the implicit reward function, we first derive part of the \method objective considering only the top-1 position from the perspective of cross-entropy. After that, we generalize the \method objective to the full image group. Finally, we justify the \method objective from the perspective of the Plackett-Luce ranking model.

\noindent\textbf{Cross entropy on top-1.}
The T2I model synthesizes groups of images \( (x_1, ...,x_k) \sim \pi(x|c)\). These images are assigned with explicit rewards \( (r_1, ...,r_k)\) from a certain type of reward model \(r\) (\textit{e.g.,} OCR), and we assume that \( r_1 \geq ...\geq r_k \).
These scores can be converted into an explicit reward distribution \( Q = \big[q(1,\tau),...,q(k,\tau)\big]\) using softmax, where \(q(i, \tau) = \text{softmax}(\frac{r_i}{\tau})\) and $\tau$ is the temperature.
This distribution \(Q\) represents the probability that item \(x_i\) should occupy the first rank (Top-1) of some ideal ranking (\textit{e.g.,} the ranking of images that generate correct texts).

Our objective is to maximize the reward of the images synthesized by the T2I model, which can be formulated using \cref{formulation: RL}. 
Therefore, given any timestep $t$, we can calculate the implicit reward function \(s_\theta(x,t)\) in \cref{formulation: diffusion_implicit_reward}. 
We can similarly apply softmax to the implicit scores $s_\theta$ calculated from \(\pi_\theta\), with \(p_\theta(i) = \text{softmax}\big(s_\theta(x_i,t)\big)\). This results in an implicit reward distribution \(P_\theta\) calculated from the diffusion model being optimized. A natural cross-entropy objective between the explicit and implicit distribution is then
\begin{align}
\mathcal{L}_{\text{top-1}}(\theta)
= \mathrm{CE}(Q,P_\theta)
= -\sum_{i=1}^{k} q(i, \tau)\,\log p_\theta(i) \notag\\
= \log\!\sum_{j=1}^{k} \exp\big( s_\theta(x_j,t)\big) \;-\; \sum_{i=1}^{k} q(i, \tau)\, s_\theta(x_i,t).
\end{align}
This objective aligns the model's top-1 implicit reward distribution $P_\theta$ with the explicit reward distribution $Q$, thereby increasing the likelihood of the high-reward samples being generated. 
The temperature $\tau$ controls the sharpness of $Q$: smaller $\tau$ amplifies reward gaps and yields a peaky distribution that concentrates mass on the highest reward items, while larger $\tau$ smooths 
$Q$, mitigating the reward differences. Note that distributions $P_\theta$ and $Q$ only focus on the probability of the top-1 position, and the scalar term of \cref{formulation: naive_implicit_reward} is also canceled (proofs in appendix)

\noindent\textbf{Cross entropy for remaining.}
To capture the full image group ranking rather than only the top-1, we view a ranking as a sequence of choices over the remaining set (\textit{e.g.,} when top-1 is fixed, we only consider the remaining $k-1$ items to get top-2). Concretely, at the choosing step \(i\), we normalize the rewards over the remaining items to obtain a target distribution
\(
    q_i(j,\tau) = \frac{exp(r_j/\tau)}{\sum_{t=i}^{k}exp(r_t / \tau)}.
\)
and define the predicted distribution
\(
    p_i(j, \theta) = \frac{exp(s_\theta(x_j)/\tau)}{\sum_{t=i}^{k}exp(s_\theta(x_t) / \tau)}
\). Following the derivations of \(\mathcal{L}_{\text{top-1}}\) and summing the cross-entropy losses across each index gives our full \method loss:
\begin{equation}
\label{formulation:soft-pl}
\small
\mathcal{L}_\text{GDRO}(\theta)
=\sum_{i=1}^{k-1}\left(
\log \sum_{m=i}^{k}e^{s_\theta(x_m, t)}
-\sum_{j=i}^{k} q_i(j,\tau)s_\theta(x_{j}, t)
\right).
\end{equation}
It's summed to $k-1$ because the last position is determined once the $k-1$ position is fixed. 
\cref{fig: pipeline} shows the overview of above calculation when $k=3$.

\noindent \textbf{Top-1 likelihood stabilization.}
During experiments, we find that though the evaluation score increases and the likelihood of the last-1 sample degrades, the likelihood of the top-1 sample also degrades (but in a slower way). This phenomenon can be explained by the theoretical justification in the following \cref{sec: Theory}. The degradation of the top-1 likelihood compromises the image quality. Therefore, we stabilize the top-1 likelihood using the following regularization:
\begin{equation}
    \mathcal{L}_\text{reg}(\theta) = M \circ || v - v_\theta\big(x_t(c), t, c\big)||^2_2,
\end{equation}
where $M$ is a one-hot mask with only the location of the top-1 sample being 1. The final objective is then 
\begin{equation}
    \mathcal{L}_\text{final}(\theta) = \mathcal{L}_\text{GDRO}(\theta) + \gamma \mathcal{L}_\text{reg}(\theta),
\end{equation}
where $\gamma$ is the regularization strength.

\subsection{Theoretical Justification from Ranking}
\label{sec: Theory}

\noindent \textbf{PL model objective.}
We can also derive the \method objective through the Plackett–Luce (PL) Model~\cite{luce1959individual, plackett1975analysis} for rankings. 
The T2I model synthesizes groups of images \( (x_1, ...,x_k) \sim \pi(x|c)\).
We assume that they are ranked as \( x_1 \succ ... \succ x_k \), but temporarily don't have access to the explicit reward function. \(x_i \succ x_j\) means that, \(x_j\) is not a better choice compared to \(x_i\) according to some criteria.
Suppose that this ranking is decided by some implicit score function \(r^*(x, c)\). Then, according to the Plackett-Luce Model, the likelihood of this ranking is
\begin{align}
    P\!\left(x_1 \succ ... \succ x_k\right)
= \prod_{i=1}^{k}
\frac{\exp\!\big(r^*(x_{i},  c)\big)}
{\sum_{j=i}^{k}\exp\!\big(r^*(x_{j},  c)\big)} \,.
\end{align}
Then, maximizing the likelihood of this distribution is minimizing the negative logarithm of this probability:
\begin{align}
\label{formulation: GDRO-naive}
\mathcal{L}
= - \sum_{i=1}^{k} \left(
 r^*\!\left( x_i, c\right)
  - \log \sum_{j=i}^{k} \exp\!\big(r^*\!\left(x_{j} \mid c\right)\big)
\right).    
\end{align}
Apply the implicit reward function
\(
r^*(x,c) = s_\theta(x,t)
\) at arbitrary $t$ to \cref{formulation: GDRO-naive}, we can get the \method loss without the constraints of explicit rewards, \textit{i.e.,} only with rankings:
\begin{align}
\label{formulation: GDRO-naive-s}
    \mathcal{L}_{\text{rank}}(\theta)
= \sum_{i=1}^{k-1} \left(
  \log \sum_{m=i}^{k} \exp\!\big(s_\theta\!\left(x_{m},t\right) \big) - \underbrace{s_\theta\!\left(x_{i},t\right)}_\text{no explicit rewards}
\right). 
\end{align}
The PL derivation above provides a ranking-centric perspective of the group-wise loss that aims at optimizing the model to follow the ranking, compared to \cref{sec: gdpo}'s cross-entropy view. 
From the ranking perspective, this loss pushes the model to enlarge the likelihood margin between the first-order sample and all remaining samples, rather than merely a collection of local pairwise wins. This perspective also explains why top-1 likelihood stabilization is necessary, since the objective aims to enlarge the likelihood margin between samples of different rankings, but does not guarantee that the likelihood of the top-1 sample increases, so an explicit objective that stabilizes the top-1 likelihood is required.

\noindent \textbf{From ranking to GDRO.}
This ranking perspective also opens a direct path to the ``soft" targets in \cref{sec: gdpo} that considers the explicit rewards. Suppose now we have access to the explicit reward, we then can replace the one-hot target at index \(i\) with a soft target distribution over the remaining samples \(\{x_i, ...,x_k\}\). Concretely, \(s_\theta(x_i,t)\) is replaced by its soft counterpart
\(\sum_{j=i}^{k}q_i(j, \tau)s_\theta(x_j, t)\), where \(q_i(j, \tau)\) is computed from the explicit reward (defined in \cref{sec: gdpo}) as a temperature(\(\tau\))-scaled softmax over the remaining items. Stacking over all steps immediately yields our \method objective \(\mathcal{L}_{\text{GDRO}}\) in \cref{formulation:soft-pl}. 
This soft replacement makes the credit distributed across the remaining candidates in proportion to their explicit rewards, rather than solely based on order information.
Note that \(\mathcal{L}_{\text{GDRO}}(\theta)\) reduces to \(\mathcal{L}_{\text{rank}}(\theta)\) when \(\tau\rightarrow 0\), and reduces to the DPO objective when \(k=2, \tau\rightarrow 0\) (proofs in appendix). 
\section{Experiments}
\label{sec: Experiments}
In this section, we first provide basic experiment and dataset settings. 
Then, we clarify reward misalignment and hacking, and try to quantify and consider the hacking using a corrected score during evaluation. 
After that, we show the main results of \method and conduct comparisons. Finally, we perform ablation studies of our method.

\subsection{Experiment Settings}
\label{sec: settings}
We use FLUX.1-dev~\cite{Flux} as the pretrained diffusion model. We use LoRA~\cite{hu2022lora} and EMA for \method training with a LoRA rank of 32 and an EMA decay rate $\eta(i)$ following $\eta(i) = \min(0.001i, 0.5)$, where $i$ is the optimization step. 
The equivalent batch size is computed following $bs = \frac{512}{k}$.
During evaluation, we use an inference timestep number of 28. During evaluation, we fix the seed for all the prompts and all methods to provide fair comparisons, as well as to observe how much different methods shift the generation distribution of the original diffusion model.
All experiments are conducted on 8 A100s and $512\times512$ resolutions.

\subsection{Datasets}
We adopt the OCR task~\cite{cui2025paddleocr30technicalreport, cui2025paddleocrvlboostingmultilingualdocument} and the GenEval task~\cite{ghosh2023geneval} following the setting of Flow-GRPO. The training set of OCR contains 19653 prompts, and the test set contains 1024 prompts. And the training set of GenEval contains 50000 prompts, with 2212 prompts being the test set.
For the training sets, we pre-generate 16 images for each prompt and assign the rewards accordingly for later offline training.

\begin{figure}[t]
  \centering
  \resizebox{!}{0.4\linewidth}{
  \includegraphics{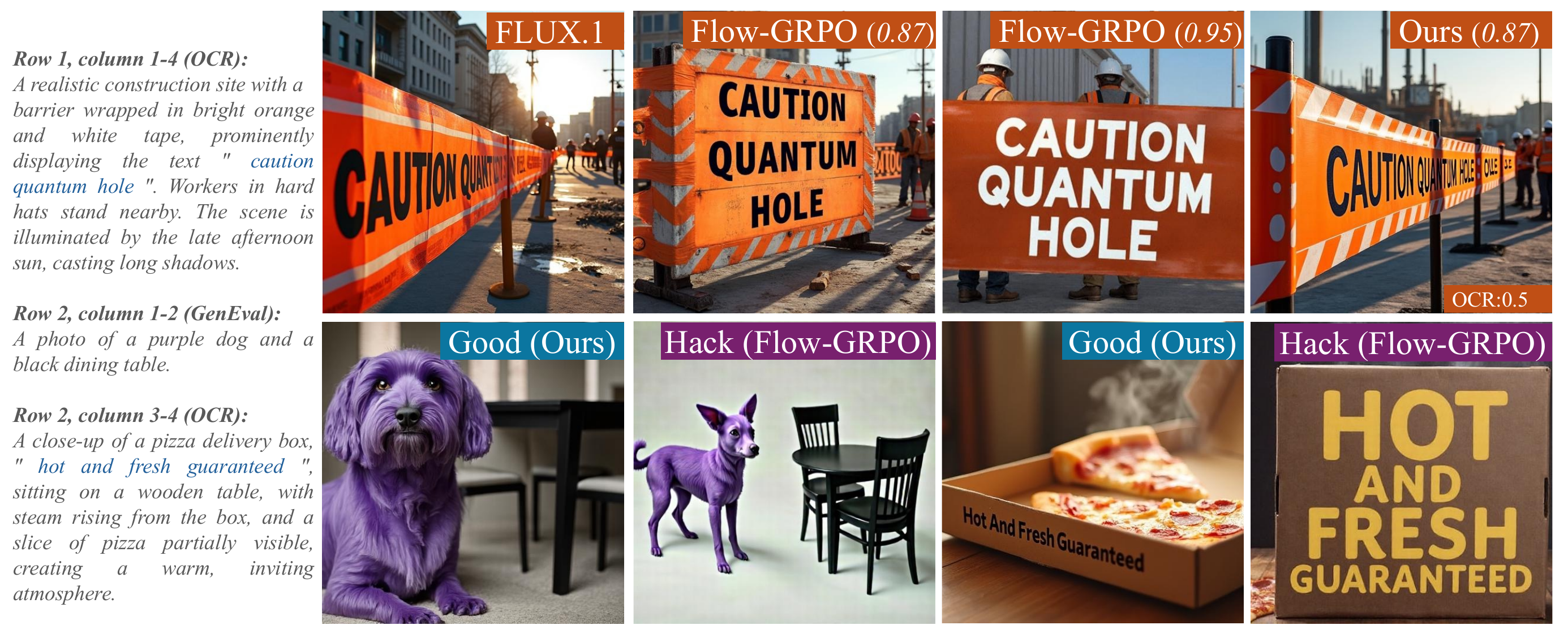}
  }
  \vspace{-20pt}
  \caption{\textbf{Reward misalignment and hacking}. 
  The first row shows the reward misalignment scenario.
  The second row shows two reward hacking cases.
  }
  \vspace{-10pt}
  \label{fig: ocr_why_hack}
  
\end{figure}

\begin{table}[t]
      \centering
      \caption{\textbf{User studies}. We let humans pick winners from different methods on OCR task to show reward misalignment and hacking.
  }
  \vspace{-10pt}
\resizebox{!}{1cm}{
  \begin{tabular}{lcccl}
    \toprule
    Method & Text Accuracy Win~($\uparrow$)  & Alignment Win~($\uparrow$) & Quality Win~($\uparrow$)\\
    \midrule
    FLUX.1 (Baseline) & 1.90\%  & 26.67\% & 33.10\% \\
    \midrule
    \textbf{Ours} (0.87)  & 5.48\%  & 33.10\% & 33.81\% \\
    Flow-GRPO (0.87)  & 3.81\%  & 16.90\% & 17.86\% \\
    Flow-GRPO (0.95)  & 6.19\%  & 9.52\% & 8.09\% \\
    Tie (baseline excluded) & 82.62\%  & 13.81\% & 7.14\% \\
  \bottomrule
  \end{tabular}
  }
  \vspace{-5pt}
  \label{tab: userstudy_reward_hacking}
  \end{table}

  \begin{table}[t]
      \centering
      \caption{\textbf{UnifiedReward scores}. We apply UnifiedReward on the results on OCR and GenEval task to indicate reward hacking.
  }
  \vspace{-10pt}
\resizebox{!}{1.4cm}{
  \begin{tabular}{lcccl}
    \toprule
    Method & Alignment~($\uparrow$)  & Coherence~($\uparrow$) & Style~($\uparrow$)\\
    \midrule
    FLUX.1 (OCR Baseline)  & 3.2572  & 3.7617 & 3.2577 \\
    \textbf{Ours} (OCR 0.87)  & 3.2764  & 3.7433 & 3.2366 \\
    Flow-GRPO (OCR 0.87)  & 3.2882 & 3.7335 & 3.1770 \\
    Flow-GRPO (OCR 0.95, severely hacked)  &  3.1164 & 3.7140 & 3.0533 \\
    \midrule
     FLUX.1 (GenEval Baseline)  & 3.2969  & 3.6191 & 3.2376 \\
    \textbf{Ours} (GenEval 0.85)  & 3.4277 & 3.5747  & 3.1647 \\
    Flow-GRPO (GenEval 0.85)   & 3.4358 & 3.5323 & 3.1142 \\
    Flow-GRPO (GenEval 0.89, severely hacked)  & 3.4456 & 3.5055 & 3.0860 \\
  \bottomrule
  \end{tabular}
  }
  \vspace{-10pt}
  \label{tab: unifiedreward_reward_hacking}
  \end{table}

\begin{figure*}[t]
  \centering
  \resizebox{!}{0.8\linewidth}{
  \includegraphics{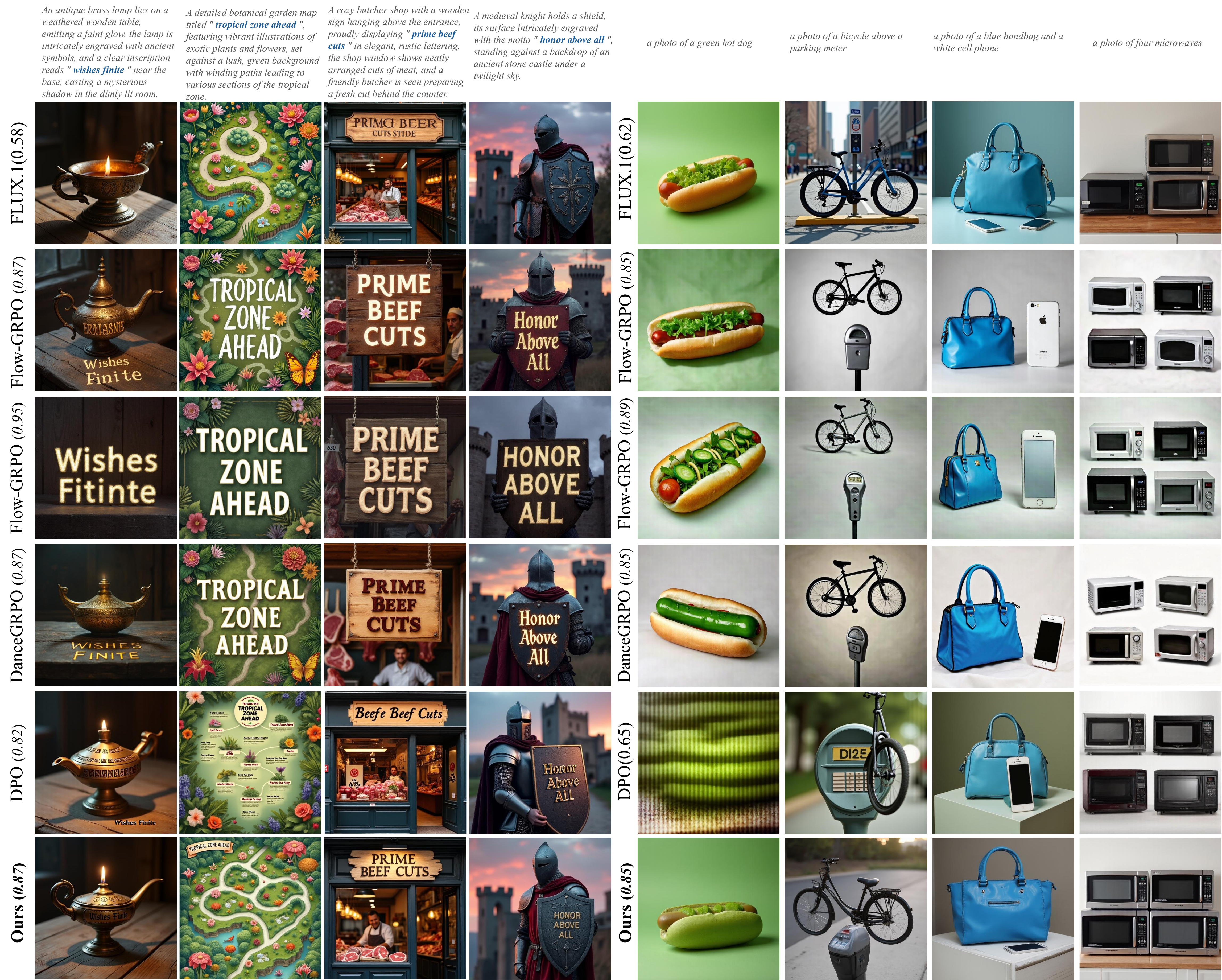}
  }
  \vspace{-20pt}
  \caption{\textbf{Qualitative results on OCR and GenEval}. Columns 1-4 show the OCR task, and Columns 5-8 display the GenEval task.
  }
  \vspace{-10pt}
  \label{fig: main_results}
\end{figure*}

\subsection{Reward Hacking and Corrected Score}
\label{sec: hacking_and_corrected}

Before analyzing the main experiments, we first clarify the existence of reward hacking by qualitative and quantitative analysis. 
Then, we model the reward hacking trend through UnifiedReward.
Finally, we propose a corrected score for more objective evaluations that considers reward hacking.

\noindent \textbf{Qualitative perspective.}
We first give examples on reward misalignment and hacking in \cref{fig: ocr_why_hack}. 
In the first row, the original FLUX.1 generates slanted and incorrect texts. 
With \method (the rightmost column), the model generates texts that are correct to humans but are too slanted to be recognized by the OCR model, resulting in a lower score of 0.50 (the full score is 1.0).
In this case, the reward is \textit{misaligned} with human perceptions.
Flow-GRPO pushes the model to enlarge and rectify the text orientation. This strategy makes the synthesized text more readable to the OCR model and thereby raises the evaluation reward to 1.0, but it makes the text overly dominate the image, which leads to detail loss and prompt-image misalignment. 
These cases demonstrate one reason for reward hacking: the reward is misaligned with human perception.
The second row shows two pairs of reward hacking cases from GenEval (left 2 columns) and OCR (right 2 columns).
All these images get full marks on each task, but the image quality shows great distinctions: the hacking cases lose most details. More pairs are in \cref{fig: teaser}.

\noindent \textbf{Quantitative perspective.}
We further study reward hacking using user studies and an automated evaluation model. 

For the user study, we chose the original FLUX.1, different Flow-GRPO checkpoints, and our method.
We randomly pick 20 sets of images for each method of the OCR task and send them to 21 humans to pick a winner for each image group (4 images) from three perspectives, including text accuracy, prompt-image alignment, and general quality. 
We also set a ``tie" option when a tie happens for methods excluding the original FLUX.1. 
The user study result in \cref{tab: userstudy_reward_hacking} shows that, in terms of text accuracy, humans mostly vote for a tie because the actual text accuracies of the three optimization methods are almost the same, though assigned with different OCR scores. 
However, \method and the FLUX.1 baseline significantly surpasses the Flow-GRPO methods and almost ties with each other from the alignment and quality perspectives. 
This shows that Flow-GRPO has already suffered from alignment and quality degradation, indicating that reward hacking truly occurs for Flow-GRPO.

Now that we know that reward hacking exists for Flow-GRPO, we then apply UnifiedReward~\cite{wang2025unified, unifiedreward-think} to measure the prompt-image alignment score, the coherence score, and the style score (ranging from 1 to 5), hoping to reflect the reward hacking from these metrics.
The results are in \cref{tab: unifiedreward_reward_hacking}.
All Flow-GRPO checkpoints exhibit severe score degradation compared to the FLUX.1 baseline across the three metrics of the two tasks (except for the alignment score of GenEval), especially for the ones with higher evaluation rewards. 
This shows that, though the model is optimized toward higher rewards, the general image quality degrades, indicating reward hacking. 
For the GenEval score increase during hacking, it is because the GenEval prompts only describe the object attributes without other descriptions of details, so the detail loss cannot be reflected by the prompt-image alignment.
This makes the prompt-image alignment score highly positively correlated with the GenEval reward and, therefore, positively correlated with reward hacking level.
\method only has a subtle coherence and style score loss, demonstrating \method's robustness in mitigating reward hacking as we have seen in the qualitative and user studies.
The above analysis demonstrates that larger evaluation rewards don't necessarily reflect better performance, and that the same evaluation rewards don't imply the same quality.

\noindent \textbf{Corrected score.}
We observe a negative correlation between UnifiedReward score and reward hacking, so this score can reflect the trend of reward hacking, though it cannot model the hacking precisely.
Therefore, we propose a corrected score to combine the evaluation reward $r$ and UnifiedReward for a more objective evaluation.
Specifically, for the OCR task, we average the three UnifiedReward scores into $\hat{u}$ and calculate the corrected score following $r_\text{corrected} = r(\hat{u}-3)+0.2$, since $\hat{u}$ falls between 3 and 4 mostly, and $0.2$ is added for convenience in drawing line charts. When reward hacking occurs, $\hat{u}$ decreases and thereby lowers the corrected score.
For the GenEval task, we only average the coherence and style score for computation, because in this case, the alignment score is positively correlated with the degree of reward hacking.

\subsection{Comparisons}
\label{sec: comparisons}

\begin{figure*}[t]
  \centering
  \resizebox{!}{0.275\linewidth}{
  \includegraphics{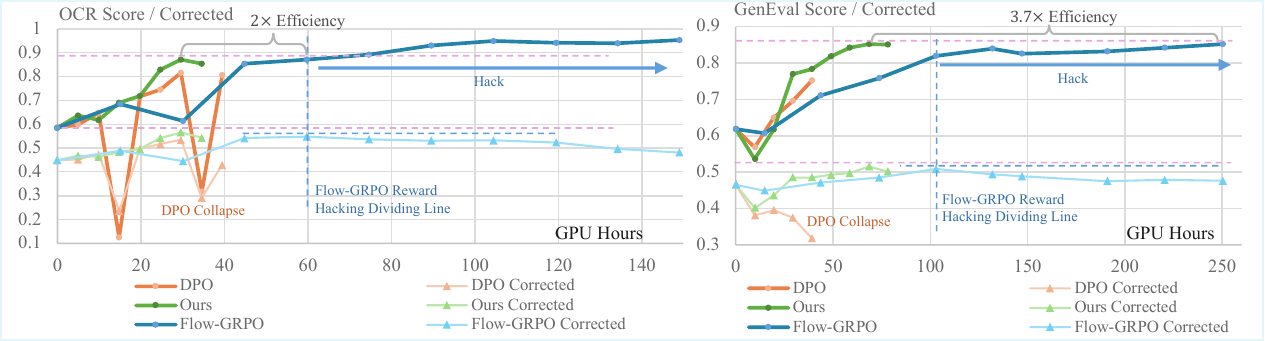}
  }
  \vspace{-22pt}
  \caption{\textbf{Evaluation curves}. We plot the evaluation time scores and corrected scores of different methods on the OCR and GenEval task across GPU training hours. We mark the reward hacking dividing line of Flow-GRPO on its peak corrected scores.
  }
  \vspace{-7pt}
  \label{fig: main_curve}
\end{figure*}

\begin{table*}[t]
      \centering
      \caption{\textbf{Quantitative results}. 
      We report the quantitative metrics on evaluation time rewards, UnifiedReward scores, and GPU hours.
    }
    \vspace{-8pt}
    \resizebox{!}{2.17cm}{
    \begin{tabular}{lccccccl}
    \toprule
    Method  & OCR / Corrected & GenEval / Corrected & Alignment Score ($\uparrow$) & Coherence Score ($\uparrow$) & Style Score ($\uparrow$) & GPU Hour \\
    \midrule
    FLUX.1 (OCR Baseline) & 0.5843 / 0.4486 & - & 3.2572  & 3.7617 & 3.2577 & - \\
     Flow-GRPO (OCR Step 160) &  0.8714 / 0.5482 & - & 3.2882 & 3.7335 & 3.1770 & 59.73 \\
     Flow-GRPO (OCR Step 400)& 0.9540 / 0.4810 & - & 3.1164 & 3.7140 & 3.0533 & 149.07 \\
     DanceGRPO (OCR Step 200)& 0.8719 / 0.5406 & - & 3.2754 & 3.7261 & 3.1704 & 74.67 \\
     DPO (OCR Step 300, Around Collapse) &  0.8158 / 0.5341 & - &  3.2735 & 3.7327 & 3.2225 & \textit{collapse} \\
     \textbf{Ours} (OCR Step 300) &  0.8721 / \textbf{0.5701} & - & 3.2764 & 3.7433 & 3.2366 & \textbf{29.60} \\
     \midrule
     FLUX.1 (GenEval Baseline) & - & 0.6178 / 0.4646 & 3.2969  & 3.6191 & 3.2376 & - \\
     Flow-GRPO (GenEval Step 680) &  - & 0.8520 / 0.4757 & 3.4358 & 3.5323 & 3.1142 & 250.27 \\
     Flow-GRPO (GenEval Step 920)& - & 0.8934 / 0.4642 & 3.4456 & 3.5055 & 3.0860 & 340.00 \\
     DanceGRPO (GenEval Step 800)& - &  0.8549 / 0.4831  & 3.4264  & 3.5391 & 3.1231 &  294.53\\
     DPO (GenEval Step 200, Around Collapse) & - & 0.6488 / 0.4162  &  3.2565 & 3.4907 & 3.1098 & \textit{collapse} \\
     \textbf{Ours} (GenEval Step 700) & - & 0.8517 / \textbf{0.5148}  & 3.4277 & 3.5747  & 3.1647  & \textbf{68.4}  \\
    \bottomrule
    \end{tabular}
    }
    \vspace{-10pt}
    \label{tab: main_table}
    \end{table*}

We compare \method with Flow-GRPO, DanceGRPO, and DPO.
For Flow-GRPO and DanceGRPO, we conduct experiments on a group size of 24 and an equivalent batch size of 576 (minimum numbers that work on FLUX.1) and $lr=3e-4$.
For DPO, when $\tau\rightarrow0$ and $k=2$, \method is equivalent to DPO, since the explicit rewards are omitted. So we perform the DPO experiments under our framework, and use $lr = 2e-4$ (higher leads to fast collapse).
For \method, 
we use the settings of $\tau=0.05, \gamma=0.5, \beta=12, lr=3e-4, k=6$ for OCR, and $\tau=0.05, \gamma=1.0, \beta=6, lr=3e-4, k=6$ for GenEval.

\noindent \textbf{Qualitative perspective.}
We present the qualitative results in \cref{fig: main_results}, with the OCR task in columns 1-4 and the GenEval task in columns 5-8. 
The first row shows images of FLUX.1 that do not meet the prompt requirements.
Flow-GRPO achieves high evaluation rewards (OCR: 0.87, GenEval: 0.85 in row 2) because the texts are more precise and easier for the OCR model to recognize, and the object attributes are more accurate. However, for the OCR task, the texts become so big that they omit many details (\textit{e.g.,}, the map in the second column turns into a giant banner, and all the map elements are lost). 
This is similar in the GenEval task, where image details almost disappear. 
These show that Flow-GRPO suffers from severe reward hacking, though the evaluation score seems high.
This reward hacking phenomenon becomes more severe when Flow-GRPO is optimized towards higher rewards (OCR: 0.95, GenEval: 0.89 in row 3), showing giant text content with severe detail losses for the OCR task, and becoming almost flat drawings for the GenEval task.
DanceGRPO shows similar behaviors.
Though DPO doesn't lose as many details as Flow-GRPO does in the OCR task (row 4, 0.82), it introduces noticeable artifacts (\textit{e.g.,} the targeted text becomes a watermark in column one and local distortions in column 3) and even fails to generate a simple case in column 3.
What's more, it tends towards collapse in the GenEval task (row 4, 0.65), as some images (\textit{e.g.,} column 1) begin to collapse.
This shows that group-level reward is important for stability.
In contrast, \method successfully optimizes toward higher scores and does not make the texts extremely big or change the image into flat drawings, making most image details preserved, and the image quality is kept. This demonstrates \method's robustness in mitigating reward hacking.

\noindent \textbf{Quantitative perspective.}
In \cref{fig: main_curve}, we plot the evaluation original and corrected scores across GPU training hours.
We first observe the GenEval curve on the right.
The DPO curve is misleading, as the GenEval score keeps increasing. However, the corrected score curve keeps dropping because more images collapse (one example in \cref{fig: main_results}) along the optimization. 
Flow-GRPO shows a slow but stable increase in the original score curve. However, its corrected score curve drops after 100 GPU hours, though the original score curve keeps increasing after that. This is because the reward hacking becomes more and more severe.
DanceGRPO is not presented here because it's a similar but slow version of Flow-GRPO.
In contrast, \method shows a fast and stable increase in the original score curve, and its peak corrected score surpasses all others, showing its efficiency and stability in optimization and mitigating reward hacking effectively. Similar phenomena can also be observed in the OCR curves on the left, except that DPO doesn't collapse as quickly as in GenEval, but it shows severe instability. \method again surpasses others in optimization efficiency and its peak corrected score.
Note that the corrected scores don't increase all the time for \method. This shows that \method cannot fully prevent all reward hacking issues, though it's much more robust toward reward hacking than others.

We present more results during evaluation in \cref{tab: main_table}. 
When optimized to the same score, \method shows $2\times$(OCR) and $3.7\times$(GenEval) efficiency compared to Flow-GRPO.
DanceGRPO takes more hours to achieve the same score as Flow-GRPO, and the corrected scores are similar due to reward hacking.
Apart from these, we also show all UnifiedReward scores to understand how they affect the corrected score. 
Though Flow-GRPO optimizes toward higher rewards, the coherence and style scores drop drastically due to reward hacking, making the corrected score decrease and lower than \method. 
A special case is the alignment score. It first increases in the OCR task for Flow-GRPO because the texts are more accurate and therefore more aligned with the prompt, but it drops later because other details are rapidly missing. 
But the alignment score increases even when reward hacking occurs in GenEval. This is because there are no descriptions of other details in GenEval prompts.
\subsection{Ablation Studies}

\begin{figure*}[t]
  \centering
  \resizebox{!}{0.3\linewidth}{
  \includegraphics{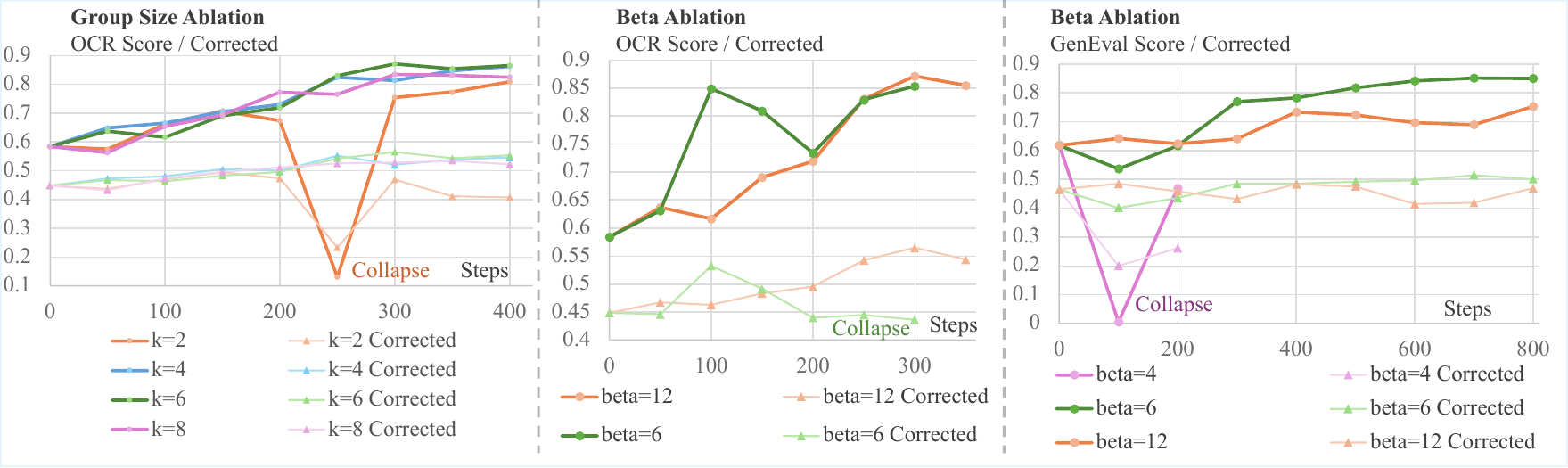}
  }
  \vspace{-25pt}
  \caption{\textbf{Curves of ablation studies.} We plot the evaluation scores across optimization steps across different ablation studies.
  }
  \vspace{-10pt}
  \label{fig: ablation_curves}
\end{figure*}

\noindent \textbf{Group size.}
We conduct ablation studies on the group size $k$ on the OCR task in \cref{fig: ablation_curves}. 
We use $\tau=0.05$, $\beta=12$, and $\gamma=0.5$.
We conduct experiments with group sizes of $ 2, 4, 6, 8$, respectively, and use a learning rate of $3e-4$.
Using a group size of 2 brings instability and collapse similar to DPO. Increasing the group size brings stability in training, as the original and corrected score curves all increase smoothly.
With a group size of 6, \method achieves the highest original and corrected score at the same time.
Except for that, we do not observe distinct differences with $k=4,6,8$. We posit that this is because increasing the group size beyond 2 brings enough stability in training.

\noindent \textbf{$\beta$ selection.} 
The $\beta$ of the implicit reward function in \cref{formulation: diffusion_implicit_reward} controls the KL regularization strength. We provide the evaluation score curves on OCR and GenEval tasks in \cref{fig: ablation_curves} using $\tau=0.05$, $k=6$, $\gamma=0.5$ for OCR, and $\tau=0.05, k=6,\gamma=1.0$ for GenEval.
For the OCR task in \cref{fig: ablation_curves}, a lower $\beta=6$ leads to a ``good" evaluation curve where the evaluation reward quickly approaches $0.85$ around step $100$. 
However, the corrected score quickly drops, demonstrating the occurrence of collapse and reward hacking.
We provide visualizations in this case in the appendix.
In contrast, a higher $\beta=12$ leads to stable and higher curves in original and corrected scores.
For the GenEval task in \cref{fig: ablation_curves}, an opposite result is observed, where the lower $\beta=6$ surpasses $\beta=12$ on both curves and doesn't bring collapse or severe reward hacking.
But setting $\beta$ even lower to $4$ leads to collapse. 
This opposite behavior on the GenEval task may be due to the reason that a bigger distribution (image layout) shift should occur to improve the GenEval score compared to the OCR score, since correcting texts requires much fewer image changes than correcting attributes, numbers, positions, and even the type of objects. Therefore, a looser $\beta$ constraint is preferred in the GenEval task. These empirical results demonstrate that a $\beta$ value should be picked carefully in different tasks.
\section{Conclusion and Discussion}
\label{sec:conclusion}

We propose \method, a post-training paradigm for diffusion models that draws the advantages of group-level reward training from LLMs. 
Derived from solid theoretical analysis, \method supports offline training while being sampler-independent, bypassing the online sampling and ODE-to-SDE conversions of RL methods, making it a method suitable for the attributes of diffusion models.
Empirical results demonstrate the effectiveness and efficiency of \method across the OCR and GenEval tasks, and the corrected score further shows that \method mitigates the reward hacking issue.

\noindent \textbf{Limitation.}
\method is currently implemented as an offline method that lacks online exploration, which may potentially limit its potential in rewards that require more active action seeking.
Also, the corrected score uses UnifiedReward to reflect the reward hacking trend, which cannot precisely capture the degree of reward hacking. A more accurate and interpretable metric can potentially be studied to better model reward hacking. 
We leave these for future studies.

\clearpage
{
    \small
    \bibliographystyle{ieeenat_fullname}
    \bibliography{main}
}

\clearpage
\setcounter{page}{1}
\appendix
\renewcommand{\thefigure}{\roman{figure}}
\renewcommand{\thetable}{\roman{table}}
\maketitlesupplementary

In the appendix, we first provide more theoretical analysis on some proofs that we omitted in the main body for simplicity in \cref{appendix: more_theory}.
Then we provide more experiment results in \cref{appendix: more_exp}, including more ablation studies in \cref{appendix: more_ablation} and more visualizations in \cref{appendix: more_visualization}.

\section{More Theoretical Analysis}
\label{appendix: more_theory}
\subsection{Implicit Reward Function}
In the main body, we introduce the implicit reward function proposed by DPO methods:
\begin{equation}
\label{appendix_formulation: naive_implicit_reward}
    s_\theta(x) = r(x_0, c) = \beta_\text{KL} \log \frac{\pi_\theta^*(x_0\mid c)}{\pi_\text{ref}(x_0\mid c)}
+ \underbrace{\beta_\text{KL} \log Z(c)}_{\text{can be deleted}}
\end{equation}
The scalar term $C = \beta \log Z(c)$ of this implicit function can be eliminated during the calculation toward the DPO objective because
\begin{small}
\begin{align}
        &\mathcal{L}_\text{DPO}(\theta) = 
        -\log\sigma\big(s_\theta(x_0^+,c)- s_\theta(x_0^-,c)\big) \notag\\
        &=-\log\sigma\big(\beta_\text{KL} \log \frac{\pi_\theta^*(x_0^+\mid c)}{\pi_\text{ref}(x_0^+\mid c)} + C -
         \beta_\text{KL} \log \frac{\pi_\theta^*(x_0^-\mid c)}{\pi_\text{ref}(x_0^-\mid c)}-C \big) \notag\\
        &=-\log\sigma\big[\beta_\text{KL} \big(\log \frac{\pi_\theta^*(x_0^+\mid c)}{\pi_\text{ref}(x_0^+\mid c)} - \log \frac{\pi_\theta^*(x_0^-\mid c)}{\pi_\text{ref}(x_0^-\mid c)}\big) \big].
\end{align}
\end{small}
This scalar term can also be canceled in the derivations toward the \method objective.
Now we prove this from the Plackett-Luce ranking model perspective.
According to the PL model, with the images $x_1,...,x_k$ and their corresponding explicit rewards $r_1,...,r_k$, the likelihood of the ranking $x_1 \succ ... \succ x_k$ is:
\begin{align}
    P\!\left(x_1 \succ ... \succ x_k\right)
= \prod_{i=1}^{k}
\frac{\exp\!\big(r(x_{i},  c)\big)}
{\sum_{j=i}^{k}\exp\!\big(r(x_{j},  c)\big)} \,.
\end{align}
The likelihood of this ranking computed from the implicit reward function is then
\begin{small}
\begin{align}
P\!\left(x_1 \succ ... \succ x_k\right)
&= \prod_{i=1}^{k}
\frac{\exp\!\big(s(x_{i},  t)\big)}
{\sum_{j=i}^{k}\exp\!\big(s(x_{j},  t\big)} \notag\\
&= \prod_{i=1}^{k}
\frac{\exp\!\big(\beta_\text{KL} \log\frac{\pi_\theta^*(x_i\mid c)}{\pi_\text{ref}(x_i\mid c)} + C\big)}
{\sum_{j=i}^{k}\exp\!\big(\beta_\text{KL} \log\frac{\pi_\theta^*(x_j\mid c)}{\pi_\text{ref}(x_j\mid c)} + C\big)} \notag\\
&= \prod_{i=1}^{k}
\frac{\exp(C)\exp\!\big(\beta_\text{KL} \log\frac{\pi_\theta^*(x_i\mid c)}{\pi_\text{ref}(x_i\mid c)}\big)}
{\exp(C)\sum_{j=i}^{k}\exp\!\big(\beta_\text{KL} \log\frac{\pi_\theta^*(x_j\mid c)}{\pi_\text{ref}(x_j\mid c)}\big)}
\notag \\
&= \prod_{i=1}^{k}
\frac{\exp\!\big(\beta_\text{KL} \log\frac{\pi_\theta^*(x_i\mid c)}{\pi_\text{ref}(x_i\mid c)}\big)}
{\sum_{j=i}^{k}\exp\!\big(\beta_\text{KL} \log\frac{\pi_\theta^*(x_j\mid c)}{\pi_\text{ref}(x_j\mid c)}\big)}
\end{align}
\end{small}
The scalar term $C$ is therefore canceled before the following derivations toward the \method objective.

\subsection{GDRO and DPO}
The objective of \method is:
\begin{equation}
\small
\mathcal{L}_\text{GDRO}(\theta)
=\sum_{i=1}^{k-1}\left(
\log \sum_{m=i}^{k}\exp\!\big(s_\theta(x_m, t) \big)
-\sum_{j=i}^{k} q_i(j,\tau)s_\theta(x_{j}, t)
\right).
\end{equation}
When $\tau\rightarrow0$, $q_i(j,\tau) = 1$ if $i=j$, and $q_i(j,\tau) = 0$ if $i\neq j$.
Therefore, when $k=2, \tau\rightarrow0$ the objective becomes:
\begin{small}
\begin{align}
\lim_{\tau\rightarrow 0}\mathcal{L}^{k=2}_\text{GDRO}(\theta) 
&=\sum_{i=1}^{2-1}\left(
\log \sum_{m=i}^{2}\exp\!\big(s_m \big)
-\sum_{j=i}^{2} q_i(j,\tau)s_j
\right) \notag \\
&=\log \sum_{m=1}^{2}\exp\!\big(s_m \big)
-\sum_{j=1}^{2} q_i(j,\tau)s_j \notag\\
&=\log\big(e^{s_1}+e^{s_2}\big)-s_1 \notag\\
&=\log\big(e^{s_1}(1+e^{s_2-s_1})\big)-s_1 \notag\\
&=\log\big(1+e^{-\Delta} \big),
\end{align}
\end{small}
where $s_i = s_\theta(x_i, t)$, $\Delta = s_1 - s_2$.

Given that $\sigma(x) = \frac{1}{1+e^{-x}}$, we can reformulate the above equation into:
\begin{align}
    \lim_{\tau\rightarrow 0}\mathcal{L}^{k=2}_\text{GDRO}(\theta) &=\log\big(1+e^{-\Delta} \big) \notag \\
    &=-\log\big(\frac{1}{1+e^{-\Delta}} \big) \notag\\
    &= -\log\sigma(\Delta) \notag \\
    &=-\log\sigma\big(s_1-s_2\big) = \mathcal{L}_\text{DPO}.
\end{align}
This means that, when $k=2, \tau\rightarrow0$, \method reduces to DPO. This is because when $\tau\rightarrow0$, the explicit reward is omitted during computation, making only the ranking matter. And when $k=2$, the ranking turns into only pair-wise preference.
From this perspective, we can view \method as a natural extension of DPO-based methods that not only enlarge the group size from 2 to any $k$, but also consider the information of the explicit rewards instead of preferences.

\section{More Experiments}
\label{appendix: more_exp}

\begin{figure}[t]
  \centering
  \resizebox{!}{1\linewidth}{
  \includegraphics{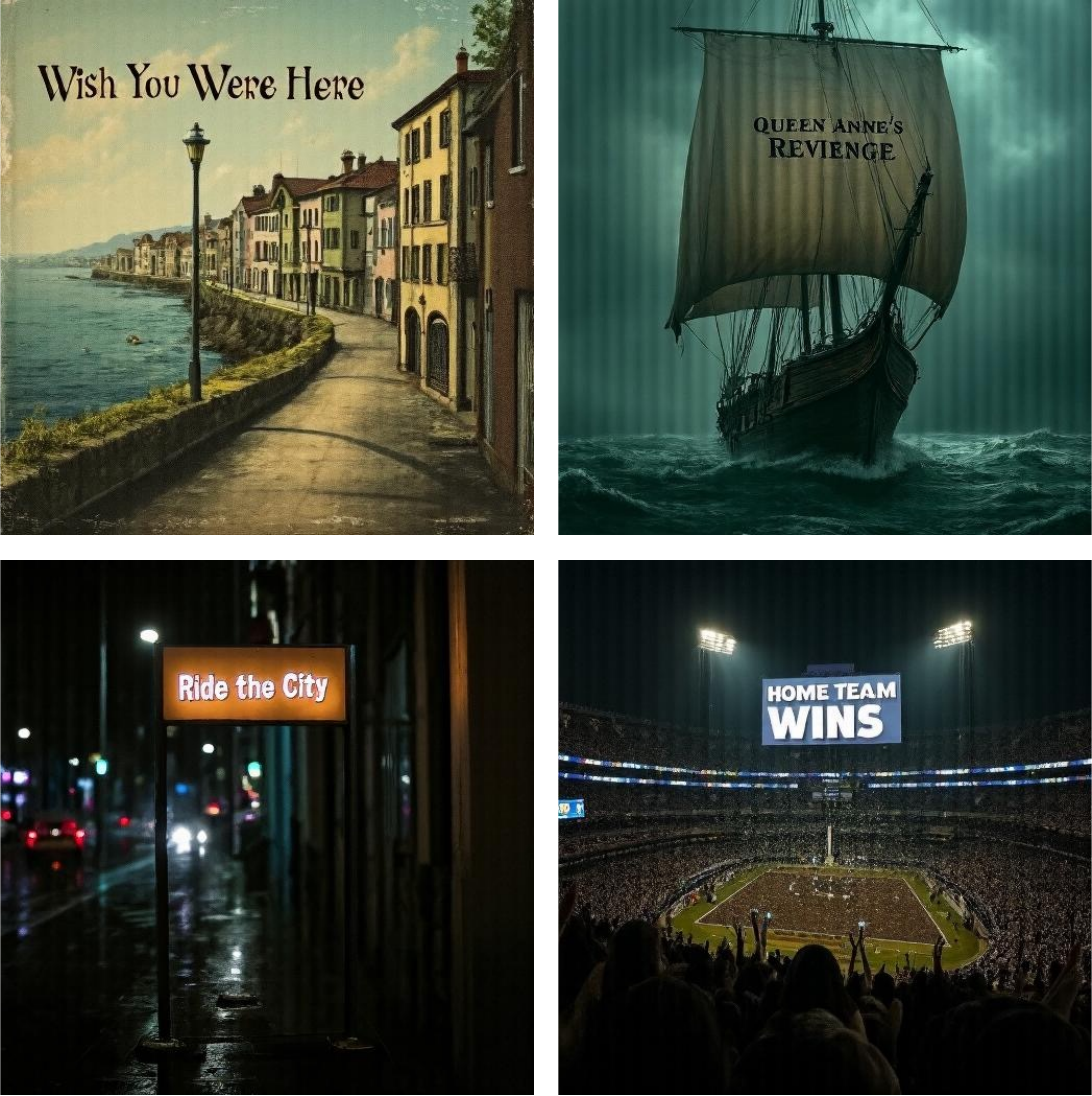}
  }
  \vspace{-20pt}
  \caption{\textbf{Demosntration on collapse}. When $\beta=6$, though an evaluation time reward of 0.85 on OCR is achieved, the images actually collapse, which is another case of reward hacking.} 
  \label{fig: appendix_collapse}
\end{figure}

\begin{figure}[t]
  \centering
  \resizebox{!}{0.6\linewidth}{
  \includegraphics{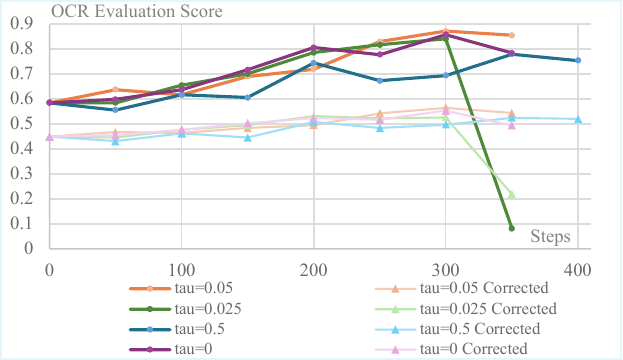}
  }
  \vspace{-20pt}
  \caption{\textbf{Ablation study on temperature}. We provide the evaluation curves on different choices of the temperature $\tau$.} 
  \label{fig: appendix_tau}
\end{figure}

\subsection{Ablation Studies}
\label{appendix: more_ablation}
\noindent \textbf{Beta $\beta$.}
We have provided the ablation study on beta in the main body. Now we provide visualizations on the case where $\beta=6$ in the OCR task in \cref{fig: appendix_collapse}. 
In the main body, the evaluation curve of $\beta=6$ in OCR is deceptive, as the original reward gets very high at an early optimization step. 
However, we show that a collapse happens for this case using the corrected score. 
Now we present the images showing what collapsed images look like, even if they earn a high evaluation time reward.
In \cref{fig: appendix_collapse}, these are evaluation images of $\beta=6$ when it achieves an evaluation reward of $0.85$ in the OCR task at step $100$. Though the number looks promising, the images demonstrate noticeable stripe artifacts as well as severe quality degradation. In this case, though the texts look right, the generation quality of the model is greatly compromised, making this optimization a failure. This further warns us that only looking at the evaluation reward is improper and deceptive.

Note that without top-1 likelihood stabilization, the optimization process will have similar collapse issues like these images, demonstrating a high reward but degraded quality.

\noindent \textbf{Temperature $\tau$.}
The temperature $\tau$ is a hyperparameter used to compute the explicit reward distribution \(q(i, \tau) = \text{softmax}(\frac{r_i}{\tau})\). This temperature controls how peaky this distribution is.
When $\tau$ is bigger, the distribution is more even and the differences between different reward values are reduced. 
When $\tau$ is smaller, the differences of the reward values are amplified during calculation.
We perform the ablation study on the choice of $\tau$ on the OCR task. With the chosen values of $0.5, 0.05,0.025$ and $0$.
Note that for the $\tau=0$ here, the explicit rewards are omitted and the explicit reward distribution reduces to a one-hot vector.
For other hyperparameters, we use $\gamma=0.5, lr=3e-4,\beta=12,k=8$.
The evaluation curves regarding $\tau$ are shown in \cref{fig: appendix_tau}.
A higher temperature $\tau=0.5$ leads to insufficient optimization, as both the original scores and the corrected score curves are all below others.
We posit that this is because a higher temperature lowers the differences between different rewards, making the model not sufficiently learns from the advantages of high-reward samples.
A lower temperature $\tau=0.025$ leads to unstable optimization, as the curves collapse.
A medium choice of temperature $\tau=0.05$ leads to the most satisfying result, demonstrating a good original reward as well as a good corrected curve, showing the best stability.
A special case is where $\tau=0$, which stands for the case without explicit reward.
In this case, \method gets a nice performance between $\tau=0.05$ and $\tau=0.025$, showing that the ranking itself can provide enough information to let the model learn the distinctions between good samples and bad samples. But still, providing more information using explicit rewards can bring more improvements.

\subsection{More Visualizations}
\label{appendix: more_visualization}

\begin{figure*}[t]
  \centering
  \resizebox{!}{1.2\linewidth}{
  \includegraphics{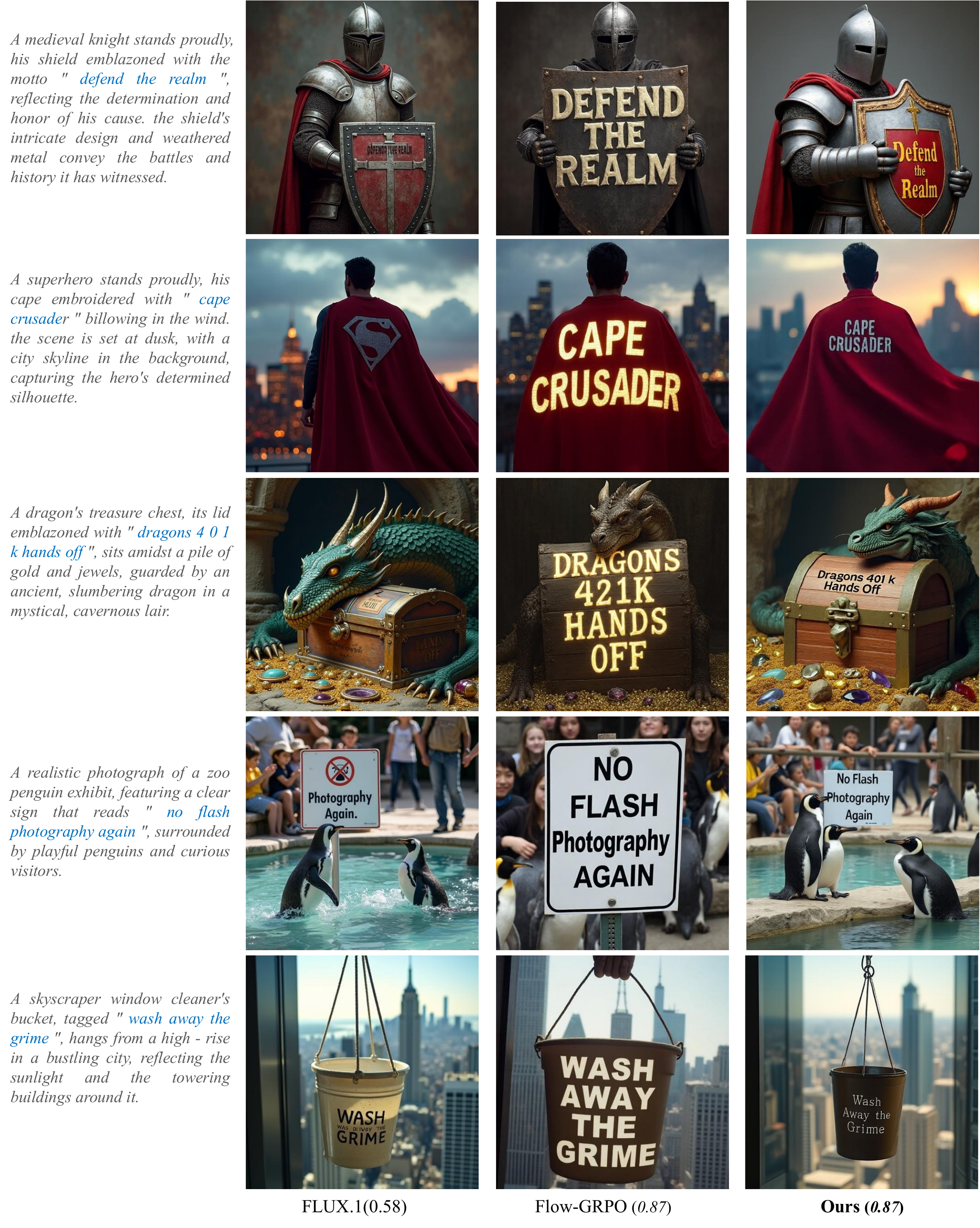}
  }
  \vspace{-10pt}
  \caption{\textbf{More visualizations on OCR}. We provide more comparisons between our method and Flow-GRPO when the evaluation reward is the same on the OCR task.
  }
  \label{fig: appendix_1}
\end{figure*}

\begin{figure*}[t]
  \centering
  \resizebox{!}{1.2\linewidth}{
  \includegraphics{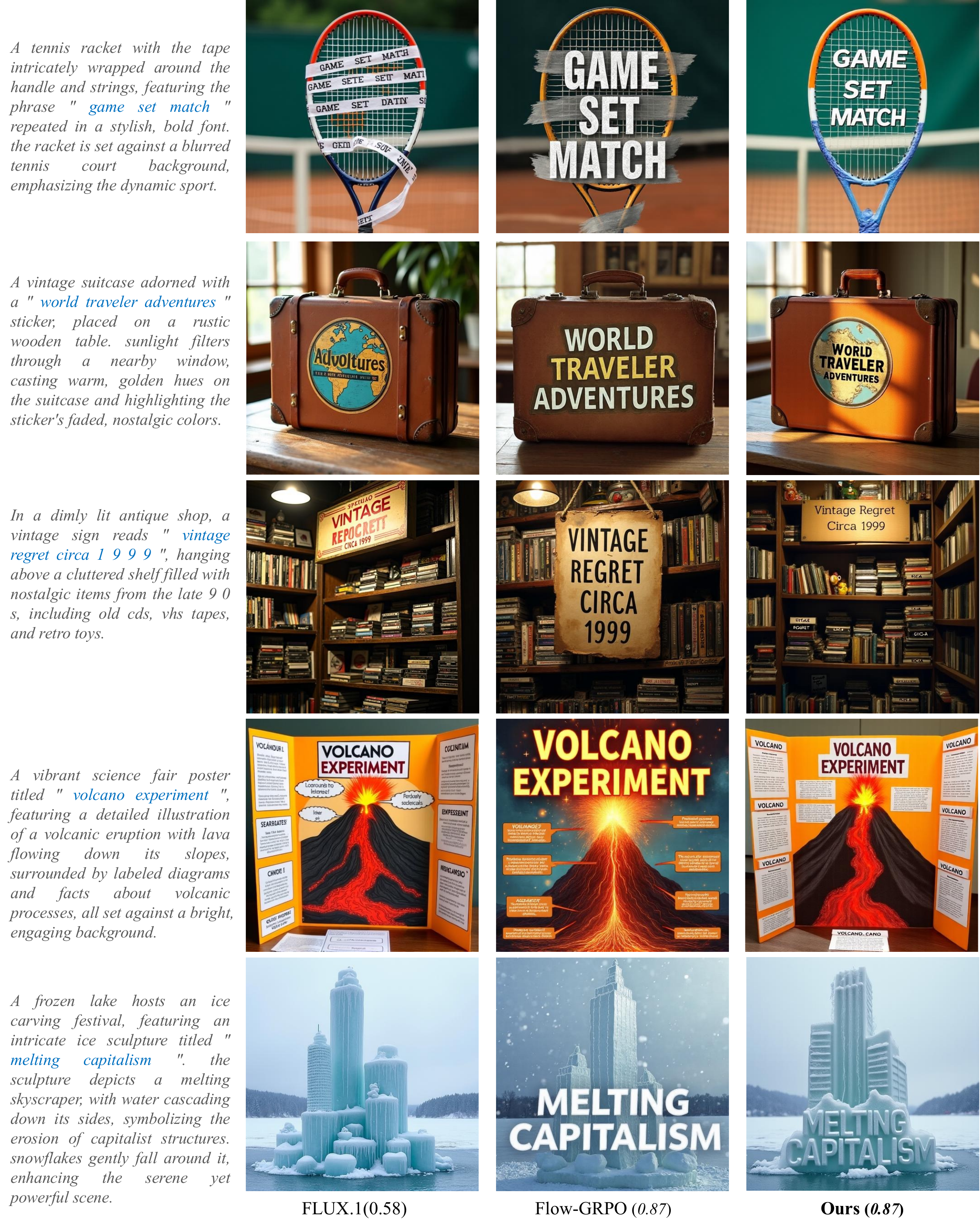}
  }
  \vspace{-10pt}
  \caption{\textbf{More visualizations on OCR}. We provide more comparisons between our method and Flow-GRPO when the evaluation reward is the same on the OCR task.
  }
  \label{fig: appendix_2}
\end{figure*}

\begin{figure*}[t]
  \centering
  \resizebox{!}{1.2\linewidth}{
  \includegraphics{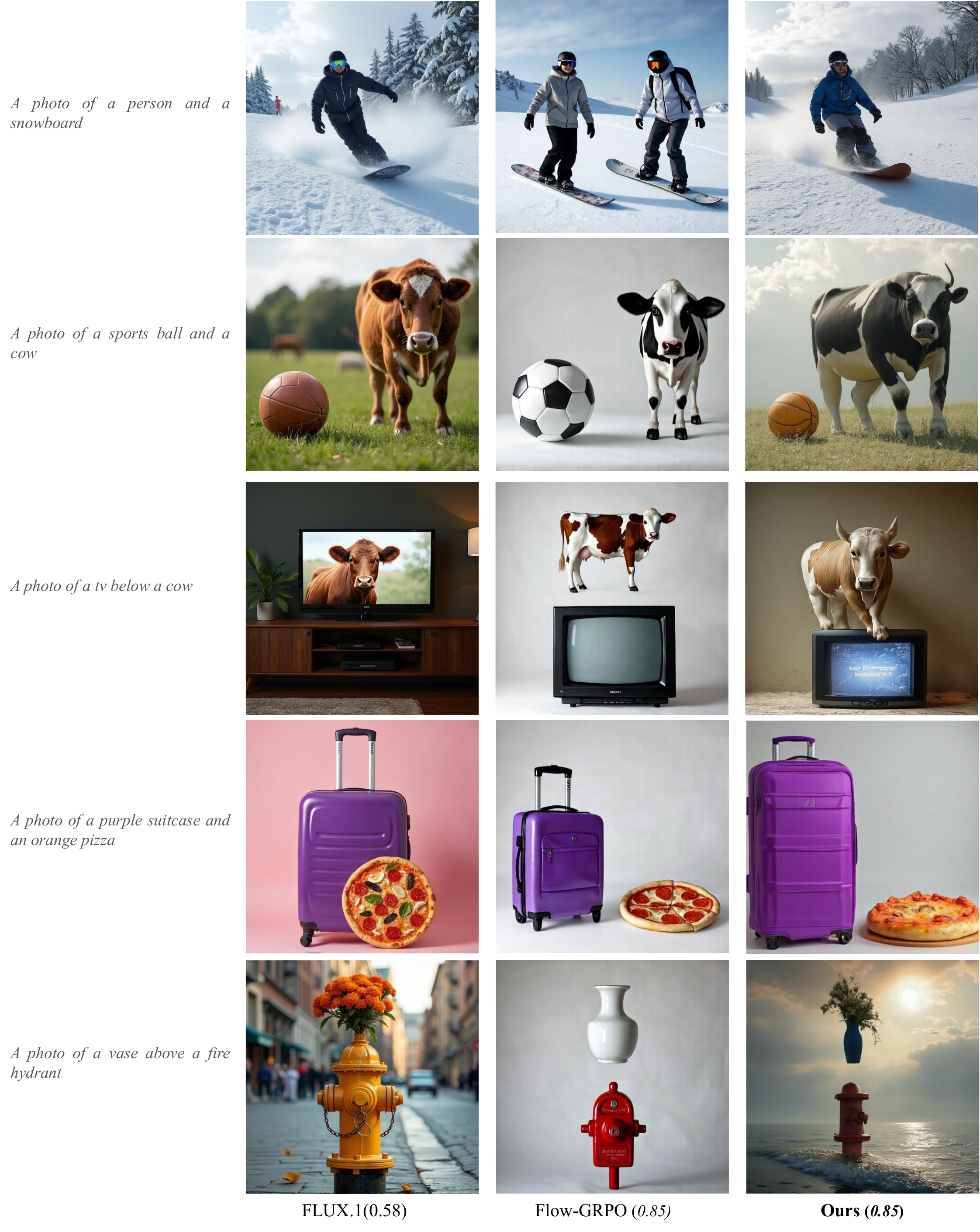}
  }
  \vspace{-10pt}
  \caption{\textbf{More visualizations on GenEval}. We provide more comparisons between our method and Flow-GRPO when the evaluation reward is the same on the GenEval task.
  }
  \label{fig: appendix_3}
\end{figure*}

\begin{figure*}[t]
  \centering
  \resizebox{!}{1.2\linewidth}{
  \includegraphics{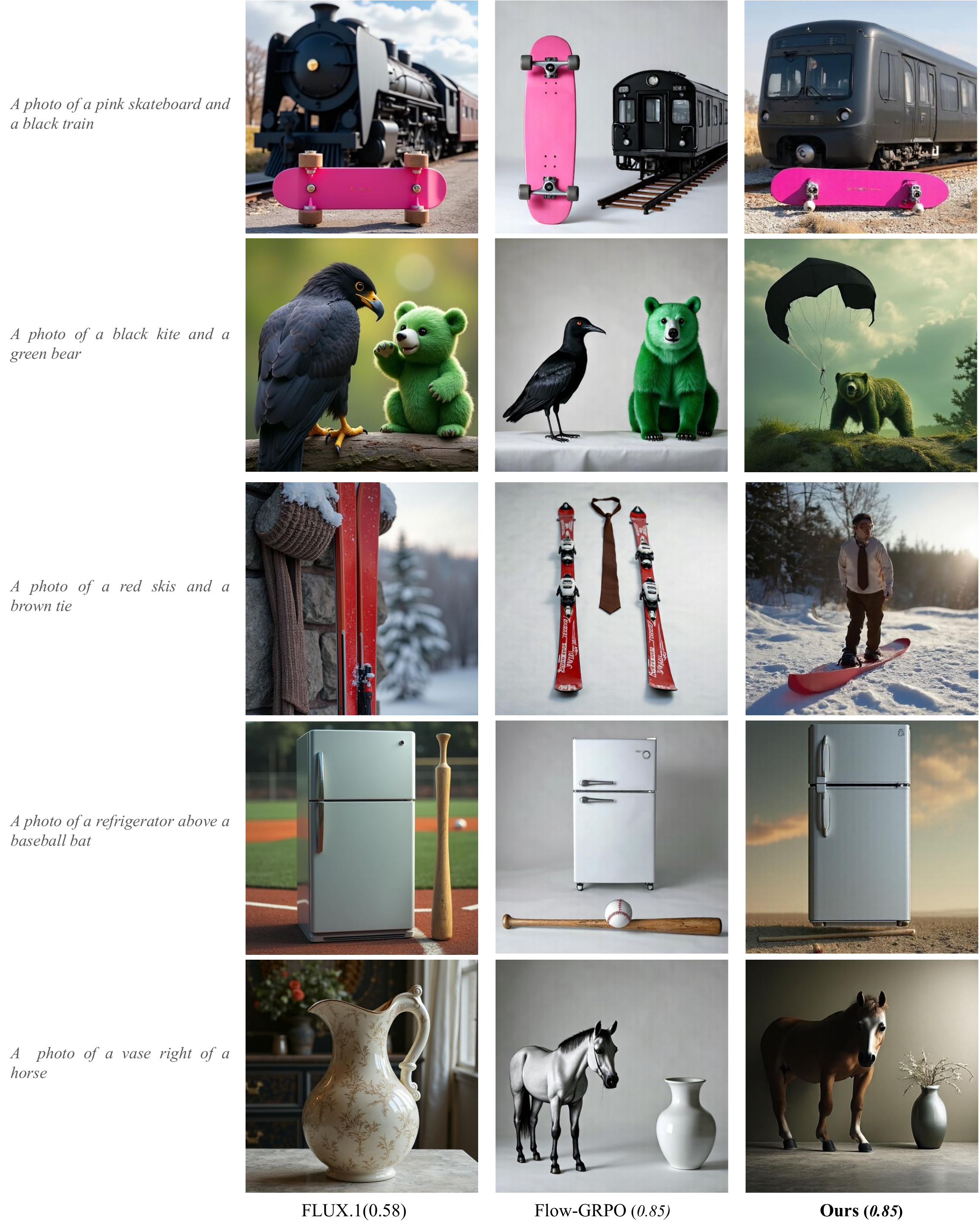}
  }
  \vspace{-10pt}
  \caption{\textbf{More visualizations on GenEval}. We provide more comparisons between our method and Flow-GRPO when the evaluation reward is the same on the GenEval task.
  }
  \label{fig: appendix_4}
\end{figure*}

We provide more visualizations of \method and Flow-GRPO.
In \cref{fig: appendix_1}, \cref{fig: appendix_2}, we provide more cases on the OCR task.
Similar to the analysis in the main body, Flow-GRPO tends to enlarge, bold, and rectify the text orientations to make the text more readable to the OCR model, and thus making other elements get omitted or even disappear. This reward hacking case is not the same as the collapse one in \cref{fig: appendix_collapse} that directly degrades the quality because of collapse, but gradually makes the details of the image fewer and thereby lowering the overall quality.
In contrast, \method doesn't have such a trend, making the images show correct texts while being abundant in details as described by the text prompts.
In \cref{fig: appendix_3} and \cref{fig: appendix_4}, we provide more cases on the GenEval task. As shown in the image, Flow-GRPO images mostly reduce to flat-drawing-like images with a plain background and disproportionate objects, ignoring most of the details. 
In contrast, \method demonstrates good visual quality as well as correct object attributes, showing its robustness in mitigating reward hacking.

\end{document}